\documentclass[runningheads]{llncs}

\usepackage{eccv}

\usepackage{eccvabbrv}

\usepackage{booktabs}

\usepackage[accsupp]{axessibility}  

\usepackage[colorlinks,citecolor=eccvblue]{hyperref}

\usepackage{orcidlink}
\usepackage{adjustbox}
\usepackage{xcolor}
\usepackage{soul}
\usepackage[utf8]{inputenc} 
\usepackage{url}            
\usepackage{booktabs}       
\usepackage{nicefrac}       

\usepackage{multicol}
\usepackage{siunitx}

\usepackage{amsmath}
\usepackage{enumerate}
\usepackage{url}     
\usepackage{graphics}
\usepackage{color}
\usepackage{multirow}
\usepackage{adjustbox}
\usepackage{bm}
\usepackage{algorithm} 
\usepackage{algpseudocode}
\usepackage{wrapfig}
\usepackage{fancybox}
\usepackage{tikz}
\usepackage{subcaption,dcolumn}
\usepackage{graphicx}
\usepackage{adjustbox}
\usepackage{tabularx}
\usepackage{diagbox}
\usepackage{makecell}
\usepackage{caption}
\usepackage{subcaption}
\newcommand\mybox[2][]{\tikz[overlay]\node[fill=gray!20,inner sep=2pt, anchor=text, rectangle, rounded corners=1mm,#1] {#2};\phantom{#2}}

\definecolor{superlightgray}{gray}{0.92}
\definecolor{lightgray}{gray}{0.8}

\newcommand{\plusplus}{\raisebox{.4\height}{\scalebox{.6}{++\xspace}}}

\newcommand{\idbox}[1]{\mybox[fill=gray!15]{\small\texttt{#1}}}

\usepackage{siunitx}
\usepackage{xspace}
\newcommand{\methodname}{\textsc{SuperGaussian}\xspace}

\begin{document}

\title{\methodname: Repurposing Video Models\\for 3D Super Resolution} 

\titlerunning{\methodname}

\author{Yuan Shen\inst{1,2}\orcidlink{0009-0001-4787-9644}\thanks{This project was done during Yuan's internship at Adobe Research.} 
\and Duygu Ceylan\inst{2}\orcidlink{0000-0002-2307-9052} 
\and Paul Guerrero\inst{2}\orcidlink{0000-0002-7568-2849} 
\and Zexiang Xu\inst{2}\orcidlink{0000-0002-8487-8018} 
\and \\ Niloy J. Mitra\inst{2, 3}\orcidlink{0000-0002-2597-0914} 
\and Shenlong Wang\inst{1}\orcidlink{0000-0002-7984-266X} 
\and Anna Frühstück\inst{2}\orcidlink{0000-0002-3870-4850}}

\authorrunning{Y.~Shen et al.}

\institute{University of Illinois at Urbana-Champaign 
\and Adobe Research 
\and University College London}

\maketitle
\def\Sr/{Super-resolution}
\def\SR/{Super-Resolution}
\def\sr/{super-resolution}
\begin{abstract}
We present a simple, modular, and generic method that upsamples coarse 3D models by adding geometric and appearance details. While generative 3D models now exist, they do not yet match the quality of their counterparts in image and video domains. 
We demonstrate that it is possible to directly repurpose existing (pre-trained) video models for 3D super-resolution and thus sidestep the problem of the shortage of large repositories of high-quality 3D training models. 
We describe how to repurpose video upsampling models -- which are not 3D consistent -- and combine them with 3D consolidation to produce 3D-consistent results. 
As output, we produce high-quality Gaussian Splat models, which are object-centric and effective. 
Our method is category-agnostic and can be easily incorporated into existing 3D workflows. 
We evaluate our proposed \methodname on a variety of 3D inputs,  which are diverse both in terms of complexity and representation (e.g., Gaussian Splats or NeRFs), and demonstrate that our simple method significantly improves the fidelity of current generative 3D models. \\ Check our project website for details: \href{https://supergaussian.github.io}{supergaussian.github.io}. 
\end{abstract}

\section{Introduction}

Generative 3D models have become a reality. Multiple methods~\cite{Anciukevivcius2023Renderdiffusion,Li2023Instant3D,Karnewar2023Holodiffusion,Shen2022SGAM} have been developed to generate 3D models, optionally conditioned using text prompts or images. These methods use a combination of image and 3D data for supervision, are fast, and produce diverse results.

Unfortunately, the 3D models generated by the current methods still lack the level of detail and accuracy achieved by state-of-the-art generative models for images~\cite{rombach2022High, Karras2019StyleGAN} or videos~\cite{Chai2023Stablevideo}. Multiple challenges contribute to this limitation. First is the choice of 3D representation. While grid-based models are most popular as they do not need prior knowledge of the (generated) shape, their regular structure (e.g., volume grid~\cite{Karnewar2023Holodiffusion, Liu2023One2345}, triplanes~\cite{Anciukevivcius2023Renderdiffusion, Bautista2022Gaudi}) puts a limit on the fidelity of the generation results. 
Secondly, acquiring large volumes of high-quality yet diverse 3D data continues to be difficult. While state-of-the-art image and video models are trained on several billion training samples, the most extensive 3D training datasets, at best, contain a few million objects. 

We explore how to increase the fidelity of generated 3D objects. Starting from any generic coarse 3D representation, our goal is to `upsample' the (coarse) 3D input model \textit{without} requiring category-specific training. Our main observation is that any 3D representation can be rendered from multiple viewpoints along a smooth trajectory and mapped to an intermediate generic video representation. 
Hence, it is possible to repurpose existing video models~\cite{zhou2023upscale, Chan2022BasicVSR++} to perform the 3D upsampling or super-resolution task. Such models are trained on large sets of video data, therefore providing strong priors that can be applied in general scenarios. 
The critical challenge is to ensure 3D consistency; while being temporally smooth, video models, however, are not guaranteed to be 3D consistent. 
Note that our approach of using video, unlike those using image-based priors and treating each frame independently, significantly improves the (initial) consistency across time. 

\begin{figure}[t]
  \centering
  \includegraphics[width=\linewidth]{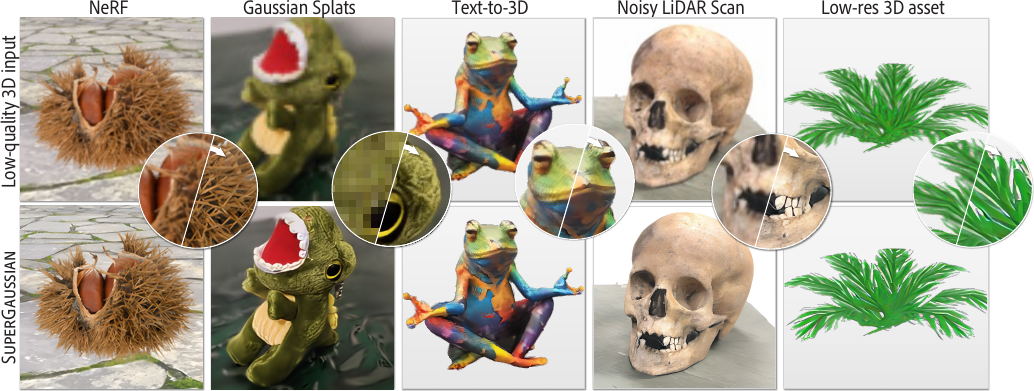}
  \vspace{-2em}
  \caption{We present \methodname, a novel method that repurposes existing video upsampling methods for the 3D superresolution task. \methodname can handle various input types such as NeRFs, Gaussian Splats, reconstructions obtained from noisy scans, models generated by recent text-to-3D methods~\cite{Li2023Instant3D}, or low-poly meshes (e.g., assets used in Sim-on-Wheels~\cite{shen2023sim}). \methodname generates high-resolution 3D outputs with rich geometric and texture details in the form of Gaussian Splats.}
  \vspace{-1.4em}
\label{fig:teaser}
\end{figure}
We address this challenge with a simple, modular, and generic approach that can be integrated into existing workflows. We proceed as follows: First, we render a video of the scene from the coarse 3D input given sampled view trajectories; second, we upsample the rendered video using a pretrained \textit{video-based upsampler} that is optionally fine-tuned to handle domain-specific artifacts of the input modality. 
For 3D consolidation, we adopt Gaussian Splatting~\cite{Kerbl2023GaussianSplats} as our output representation. Being an object-centric representation, Gaussian splats are ideally suited for encoding individual objects and are capable of capturing local details. Gaussian splats also strike a good balance between simplicity, fidelity of encoded models, and efficiency of rendering. Fig.~\ref{fig:teaser} shows that our proposed video-based 3D super-resolution framework can be successfully adapted to upscale diverse sets of coarse 3D modalities.

We evaluate our algorithm on various low-resolution assets and report the quality and diversity of the generated high-resolution Gaussian splats. We compare our algorithm against possible alternatives and report an ablation study of our design choices. 
In summary, our main contributions include: 
\begin{enumerate}[(i)]
    \item re-purposing pretrained video upsampling models for 3D super-resolution; 
    \item finetuning video upsampling prior using generic 3D data to handle artifacts characteristic to low-res Gaussian splats; and 
    \item proposing \methodname as an efficient, class-agnostic, and modular 3D super-resolution procedure that successfully upsamples diverse 3D scene representation, including Gaussian splats, NeRF, and others. 
\end{enumerate}

\section{Related Work}
\textbf{Image \SR/.}
Enhancing image details and resolution is a long-standing problem in computer vision. 
This is inherently an ill-posed problem, so generative methods are well suited to tackle the hallucination of missing information. For conciseness, we concentrate on recent generative approaches and refer the reader to a comprehensive review~\cite{Chen2022SuperResolutionSTAR} of this subject area. 

Early CNN-based approaches~\cite{Dong2014Learning, Dong2016Accelerating, Zhang2018Image, Kim2016Accurate} recognize the applicability of deep networks for super-resolution.
Several approaches leverage powerful priors learned by image synthesis models to infer image details. Generative Adversarial Networks~\cite{Goodfellow2014GANs} have been extensively used for this task, either through optimization-based techniques~\cite{Menon2020PULSE} or conditional feed-forward network designs~\cite{Ledig2017PhotoSR, Wang2018Recovering, Wang2018ESRGAN, Wang2021RealESRGAN, Bahat2020Explorable, Kang2023GigaGAN}.
Other approaches have explored normalizing flows~\cite{Lugmayr2020SRFlow} or transformer-based methods~\cite{Chen2023Activating, Chen2023HAT}. 
The immensely popular denoising diffusion probabilistic models~\cite{Ho2020Denoising, Rombach2022StableDiffusion} have also been recently explored for image upsampling~\cite{Zhang2020USRNet, Li2022SRDiff, Saharia2023SR3}.

\noindent\textbf{Video \SR/.}
Following the success of learning-based single-image super-resolution methods, there have been various efforts to extend this success to the video domain. Several strategies have been explored to make this extension. While some work~\cite{Liu2017Robust} introduces temporal aggregation modules to aggregate information across time, recurrent neural networks~\cite{Yi2019Multi,chan2021basicvsr} with uni- and bi-directional information sharing have also been a choice of architecture. Deformable convolutions~\cite{wang2019edvr,tian2020tdan} have been utilized to align per-frame features both for video upsampling and restoration tasks. More recently, diffusion-based video upsamplers have been proposed by injecting temporal layers into base diffusion model~\cite{zhou2023upscale}. In our work, we leverage VideoGigaGAN~\cite{xu2024videogigagan}, a state-of-the-art generative video upsampler. However, our method is agnostic to the video model we use and can further be improved in the future with advances in video upsampling techniques, e.g., diffusion-based video upsampler.

\noindent\textbf{Image-Based 3D \SR/.}
Several approaches have recently aimed at increasing the resolution of implicit radiance fields, in particular Neural Radiance Fields (NeRFs). This is an important and practical application for both capture and synthesis scenarios where higher detail and quality are desirable. 
Starting with a low-res radiance field, at each iteration, this thread of work~\cite{Han2023SuperNeRF,Huang2023RefSRNeRF,Wang2022NeRFSR} renders images from the current representation. Next, a pre-trained image up-sampler is used to upsample the renderings. Then the gradient flow coming from the difference between the current renderings and upsampled images, is used to optimize the 3D representation. Similarly, CROP~\cite{Yoon2023CrossGuidedOptimization} also leverages a single image upsampler but introduces an uncertainty volume to improve the multi-view consistency. Unlike these methods, however, we adopt a video upsampler instead of a single-image upsampling network. Thanks to better handling of temporal consistency, we show that using the video upsampler improves the visual quality of the final results. Moreover, unlike previous work mainly focusing on upsampling NeRF representations, \methodname is a generic approach, which we demonstrate on various 3D input types.

In the context of 3D-aware GANs, several works have been proposed~\cite{Chan2022EG3D,dong2023ag3d} that jointly learn to generate a triplane-based 3D representation as well as a 2D image upsampling module. The upsampling module is applicable only to the specific triplane features learned by the generator and the final output is not fully 3D consistent since each view is upsampled independently.

\noindent\textbf{Feed-forward 3D \SR/.} 
Several 3D generation and reconstruction methods have proposed direct 3D superresolution modules. Smith et al.~\cite{NIPS2018_7883} represent a 3D shape as multi-view orthographic depth maps which are then upsampled. The recent diffusion-based 3D generation method LASDiffusion~\cite{zheng2023lasdiffusion} presents a method to predict signed distance fields from an occupancy map to improve the resolution of the 3D shapes. Both methods, however, focus only on geometry refinement while our work jointly performs appearance and geometry upsampling. Rodin~\cite{Wang2022RODINAG}, a triplane-based diffusion method, presents a triplane upsampling strategy. This approach is applicable only to the specific triplane features learned. 
Finally, our approach does not need large-scale 3D datasets and instead leverages existing video models trained on video data.

\noindent\textbf{Deep Implicit Representations.}
Recently, neural implicit representations such as NeRF~\cite{Mildenhall2021NeRF}, VolSDF~\cite{Yariv2021VolSDF}, and NeuS~\cite{Wang2021NeuS} have been popularized as an efficient and compact representation for 3D shapes. Efficiently capturing high-resolution details, however, still remains a challenge for such representations due to the low-dimensional volumetric structure and the computationally intensive volumetric rendering process. 
Recently, Gaussian Splatting~\cite{Kerbl2023GaussianSplats} has been proposed as an alternative that provides an object-centric representation with fast rendering capabilities. The scene is represented by a large number of 3D Gaussian blobs whose properties are optimized through differentiable rendering. In our work, we adapt Gaussian splats as the final 3D representation output due to the efficient optimization and rendering capabilities they offer.

\section{Our Approach}
\vspace{-3pt}
\begin{figure*}[t]
  \centering
  \includegraphics[width=\linewidth]{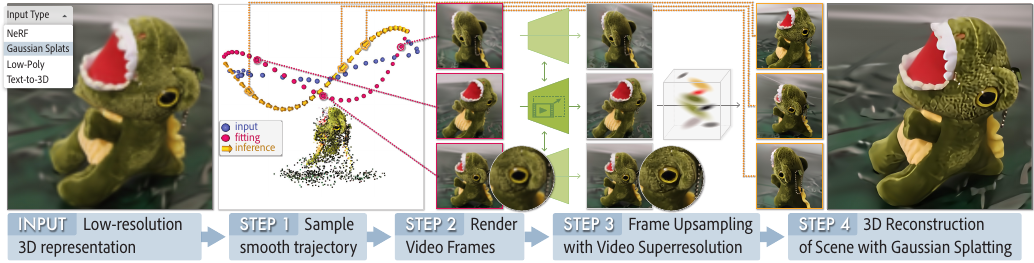}
  \vspace{-2em}
  \caption{
  \textbf{\methodname pipeline.} 
  Given an input low-res 3D representation, which can be in various formats, we first sample a smooth camera trajectory and render an intermediate low-resolution video. We then upsample this video using existing video upsamplers and obtain a higher resolution 3D representation that has sharper and more vivid details. Our method, \methodname, produces a final 3D representation in the form of high-resolution Gaussian Splats.
  }
  \vspace{-6.5mm}
\label{fig:pipeline}
\end{figure*}
\subsection{Overview}
Given a coarse 3D representation, our goal is to perform super-resolution to improve the fidelity of the 3D representation and capture more local details. Based on the observation that 3D content can be represented as a video depicting the 3D scene from multiple viewpoints, the main premise of our work is to leverage existing video upsampling priors for the task of 3D upsampling. As illustrated in Fig.~\ref{fig:pipeline}, our method \methodname consists of two main steps. First, we run upsampling on the video rendered from the coarse 3D representation to gain resolution and obtain sharp results. Then, we perform 3D reconstruction to generate a consistent 3D representation. In addition to leveraging the prior of the video upsampler, which is trained on a large set of video data, we perform fine-tuning over domain-specific low-res videos (i.e., videos rendered from low-res 3D representations). Hence, \methodname is able to handle complicated degradation caused by various 3D capture and generation processes. Note that each component in our framework is highly modularized and could easily be replaced with other state-of-the-art video methods. 

Next, we unfold this section by formulating the input and output of our method (Sec.~\ref{sec:formulation}), introducing the key video upsampling (Sec.~\ref{sec:video_upsampling}) in our framework, and discussing the details of the 3D optimization (Sec.~\ref{sec:opt}). 
\vspace{-10pt}
\subsection{Problem Formulation~\label{sec:formulation}}
\vspace{-2pt}
Our framework can handle a diverse set of coarse 3D representations of a static scene, which we denote as $\psi_{\mathit{low}}$. For example, $\psi_{\mathit{low}}$ can be Gaussian Splats, NeRF, a low-poly mesh, a low-quality captured video, or a generated 3D object with recent text-to-3D methods~\cite{Li2023Instant3D}. Any such 3D representation, $\psi_{\mathit{low}}$, can be rendered from multiple viewpoints to yield a video, a common intermediate representation. In practice, we render each 3D input from $N$ smooth trajectories with viewpoints $\xi_{1\dots T}^{1\dots N}$ resulting in a sequence of RGB images $I_{1\dots T}^{1\dots N}$. Here, the subscript indexes the viewpoint or the pose along each trajectory, and the superscript denotes the trajectory ID. 
We assume the camera movement between adjacent frames is sufficiently small such that a standard video upsampler can leverage enough temporal alignment. After we perform a sequence of video upsampling, as output, we generate a high-fidelity 3D representation $\psi_{\mathit{high}}$ in the form of Gaussian Splats (note that camera views are known in our setup and do not need to be estimated). This final 3D optimization produces a true 3D output and, in the process, removes any remaining temporal inconsistencies in the refined video representations. 

\subsection{Initial Upsampling\label{sec:video_upsampling}}
First, we manually sample trajectories near the target scene in the empty region. Given the trajectory $I_{1...T} \in \{\mathbb{R}^{W\times H \times 3}\}$ describing the camera path for each individual video, the video upsampler outputs a trajectory with $\times r$ upsampling ($r=4$ in our experiments). Mathematically,
\begin{equation}
    \hat I_{1...T} = f(I_{1...T}) \in \mathbb{R}^{rW\times rH \times 3},
\end{equation}
where $f$ denotes the video upsampler and $\hat I_{1...T}$ is the upsampled video. We assume the initial rendering resolution should be sufficiently high such that the rendered fidelity is bottlenecked by the coarse level of the input 3D representation. 

Our framework can easily integrate any of the state-of-the-art pre-trained video upsamplers. For our case, we use VideoGigaGAN~\cite{xu2024videogigagan}, a generative video upsampler. Please refer to Sec.~\ref{sec:baselines} to see how we set up the video upsampler to have fair comparisons with baselines. In cases of handling input representations with severe domain bias (see Fig.~\ref{fig:diverse_input_modality}), additional fine-tuning is desired. For example, with stripy or blob-like artifacts after zooming in, renderings from Gaussian Splats follow different degradation from standard augmentations deployed in state-of-the-art video upsamplers. Hence, to finetune the video upsampler, we need pairs of low and high-resolution videos that depict the specific degradation we would like to model. For this purpose, we use the multi-view dataset MVImgNet~\cite{yu2023mvimgnet}, which depicts a variety of 3D objects and scenes. First, we bilinearly downsample the original images in the dataset by a factor of 8, i.e., to $64\times64$px resolution, to obtain a set of low-resolution images. We then fit low-resolution Gaussian Splats to these images~\cite{Kerbl2023GaussianSplats} as described in more detail in Section~\ref{sec:opt}.  
We render the optimized low-res Gaussians in the original camera trajectory provided by the dataset as input to the video upsampler. As the target ground truth, we use the original videos from the dataset, which are resized to be 4 times the resolution of the input. We finetune the model using the Charbonnier regression loss~\cite{barron2019general} for its robustness to outlier pixels, the LPIPS loss~\cite{zhang2018perceptual} for its perceptual level improvements, and GAN loss~\cite{goodfellow2014generative} for its generative behavior. 
\vspace{-10pt}
\subsection{3D Optimization via Gaussian Splats}
\label{sec:opt}
 We use the official Gaussian Splatting codebase\footnote{\scriptsize\url{https://github.com/graphdeco-inria/gaussian-splatting}} to perform 3D optimization to fit Gaussians to the upsampled videos. In particular, for our experiments, we follow the standard Gaussian Splatting optimization process by running 2k steps. Note that as we have perfect camera information, we directly provide them to the optimization, rather than estimating them using SfM. As for our loss function, following the practice of the original paper~\cite{Kerbl2023GaussianSplats}, we use $L_1$ and $L_{\mathit{SSIM}}$ losses when optimizing the Gaussians. The advantages of adopting 3D Gaussian Splatting lie in its being an object-centric representation and its efficiency in terms of training and rendering. Besides, it is able to capture view-dependent effects well for the upsampled frames. However, \methodname can also be easily integrated with other types of 3D representations, e.g., Neural Radiance Fields, which we use in our experiments on the Blender-synthetic dataset~\cite{Mildenhall2021NeRF}.  
 
\section{Experiments}
\subsection{Setup}
\textbf{Dataset:} We finetune the video upsampler component of our method on the MV-ImageNet dataset~\cite{yu2023mvimgnet}, which consists of a rich number of real object-centric videos with high quality. This dataset is processed to fit low-resolution Gaussian Splats reconstructed from $64\times64$px posed images as described in the previous section. In total, ~200k scenes are used for training and 5k scenes for validation. A separate set of 500 scenes are used for quantitative evaluation. We obtain the ground-truth high-resolution images by center-cropping and resizing them to square images at $256\times256$px resolution on MVImgNet. To demonstrate the generalizability of our approach, we also show results on the Blender-synthetic dataset, consisting of synthetic 3D objects. We use the original data split as prior work~\cite{Wang2022NeRFSR, Lin2024FastSR} for evaluation.
\newline
\newline
\textbf{Metrics:} On the MVImgNet dataset, we utilize metrics focused on perceptual quality when performing quantitative comparisons, i.e., LPIPS~\cite{zhang2018unreasonable}, NIQE~\cite{6353522}, Inception score~\cite{salimans2016improved} and FID~\cite{heusel2017gans}. For the FID computation, we use the real images provided by MVImgNet as the real distribution. On the NeRF Synthetic dataset, we report the same metrics, i.e., LPIPS, PSNR, and SSIM, analogous to FastSR-NeRF~\cite{Lin2024FastSR} and NeRF-SR~\cite{Wang2022NeRFSR}.
Aside from quantitative metrics, we strongly encourage the readers to check our qualitative results and videos in the supplementary. 
\newline
\newline
\textbf{Baselines: \label{sec:baselines}} While there are a recent few domain-specific 3D upsampling solutions~\cite{yang2024gaussianobject, Huang2023RefSRNeRF, Wang2022NeRFSR}, we focus on comparing against baselines that are applicable to generic 3D representations. Previous work for generic 3D upsampling follows a similar paradigm by leveraging upsampling priors to iteratively optimize the chosen 3D representation. The popular choice of the prior is image-based upsampling models, instead of our proposed video-based upsampler, one seemingly trivial design choice that profoundly impacts the upsampled quality. With that, we narrow down to three state-of-the-art methods: (1) Instruct-G2G (Instruct-NeRF2NeRF~\cite{HaqueInstructNeRF2023} with Gaussian Splats as 3D format), (2) Super-NeRF~\cite{Han2023SuperNeRF} with Gaussian Splats as 3D format, and (3) FastSR-NeRF~\cite{zhao2023fast}. For completeness, we include one domain-specific baseline, NeRF-SR~\cite{Wang2022NeRFSR}. Besides, we introduce a customized baseline, \emph{Pre-hoc image}, as discussed below.

For a fair comparison, we use the same state-of-the-art image upsampler, GigaGAN~\cite{Kang2023GigaGAN} for all the baselines. Specifically for the GigaGAN checkpoint, we follow the same strategy proposed in the original paper to train a $4 \times$ super-resolution model on LAION~\cite{schuhmann2022laionb} that learns to upsample images from $64\times64$ to $256\times256$px with great details. In \methodname, we use VideoGigaGAN~\cite{xu2024videogigagan} as the video upsampler, which mostly re-uses the former GigaGAN architecture except for additional BasicVSR\plusplus~layers for temporal feature extraction and propogation~\cite{Chan2022BasicVSR++} that processes the low-res video input frames. The model size of our video prior turns out to be slightly smaller than the image
prior after adjusting feature dimensions. Finally, we finetune both the image and video upsampler on MVImgNet until convergence. Benchmark of both priors are reported in Tab.~\ref{tab:prior_type_ablation}. Both priors show generative behaviors after upsampling, and the video prior shows great temporal consistency across upsampled frames. Note \methodname can be easily integrated with a stronger video prior, e.g., Upscale-a-video~\cite{zhou2023upscale} or Sora from OpenAI~\cite{sora} by finetuning over low-res video as conditions. For Instruct-G2G, we first optimize 2k steps for the warm-up stage, which prepares the low-res representation, and then optimize another 2k steps with upsampling by upsampling every 20 steps. For SuperNeRF, since the authors did not release their codebase, we self-implemented their method with our best knowledge. We wrap our GigaGAN with the consistency enforcement module~\cite{Bahat2020Explorable}. Similarly, we optimize 2k step for low-res representation and then optimize for another 2k steps. We set the learning rate of the CCLC embedding to be \num{5e-5}, which is critical to be small to avoid blurry results. 
\begin{table}[t]
\centering
\def\arraystretch{1.28}
{
\scriptsize
\caption{{\bf Benchmark of video and image upsampler priors}. We report the performance of the upsampled frames for pre-hoc image and ours after applying the corresponding upsampler. Note: reconstruction is not performed for these results. 
\label{tab:prior_type_ablation}
}
\vspace{-2mm}
\begin{tabular}{c|cccc}
    \specialrule{.1em}{.1em}{.1em}
      {Model} & {~LPIPS $\downarrow$~} & {~NIQE $\downarrow$~} & {~FID $\downarrow$~} & {~IS $\uparrow$~}  \\ \hline
      image upsampler & 0.1533 & 7.06 & 14.67 & 11.67 {~$\pm$ 1.14}  \\
      video upsampler & \textbf{0.1511} & \textbf{6.04} & \textbf{14.61} & \bfseries 12.45 {~$\pm$ \textbf{1.19}} \\  
    \specialrule{.05em}{.05em}{.05em}
\end{tabular}
}
\vspace{-2em}
\end{table}
In our experiments, we observe that running an iterative optimization with an image upsampler, i.e., SuperNeRF~\cite{Han2023SuperNeRF} and Instruct-G2G, often results in more blurry results compared to upsampling all the frames individually first, in a \emph{pre-hoc} manner, and then performing an optimization. Hence, we define another baseline, \emph{Pre-hoc image}, where we individually upsample all the frames once and then optimize the 3D representation.  

Finally, to remove the effect of representation on the experiment results, we modify our baselines to all use 3D Gaussian Splattings as their global representation for comparisons on MVImgNet. On the Blender synthetic dataset, Neural Radiance field is chosen as the global representation across methods to faily compare with other baselines represented in NeRF. 
\newline
\newline
\textbf{Implementation details:} During finetuning, all model parameters are fine-tuned with a learning rate of \num{5e-5}. We weigh the LPIPS, Charbonnier regression losses, and adversarial loss with weights \num{15}, \num{10}, and \num{0.05}, respectively. Following GigaGAN~\cite{Kang2023GigaGAN}, we apply $R_1$ regularization~\cite{mescheder2018training} on the discriminator at a weight of 0.02048 and interval of 16. Batch size is 64 with 12 video frames in each. Our models are trained on 64 A100 GPUs, each with 80GB VRAM. 

For a set with $30$ frames, the video upsampling takes $\sim$1 second to upsample $64\times 64$ inputs. 
High-resolution Gaussian splat optimization takes around $30$ seconds for 2k steps to fit images of resolution $256\times 256$. On average, our pipeline takes 141.33 sec to complete, the most efficient compared with baselines. We note that our codebase can be further optimized for efficiency as there is substantial room for improvement in I/O and pipelining.

\vspace{-10pt}
\subsection{Comparison Studies}
\begin{figure}[ht!]
  \centering
  \includegraphics[width=\linewidth]{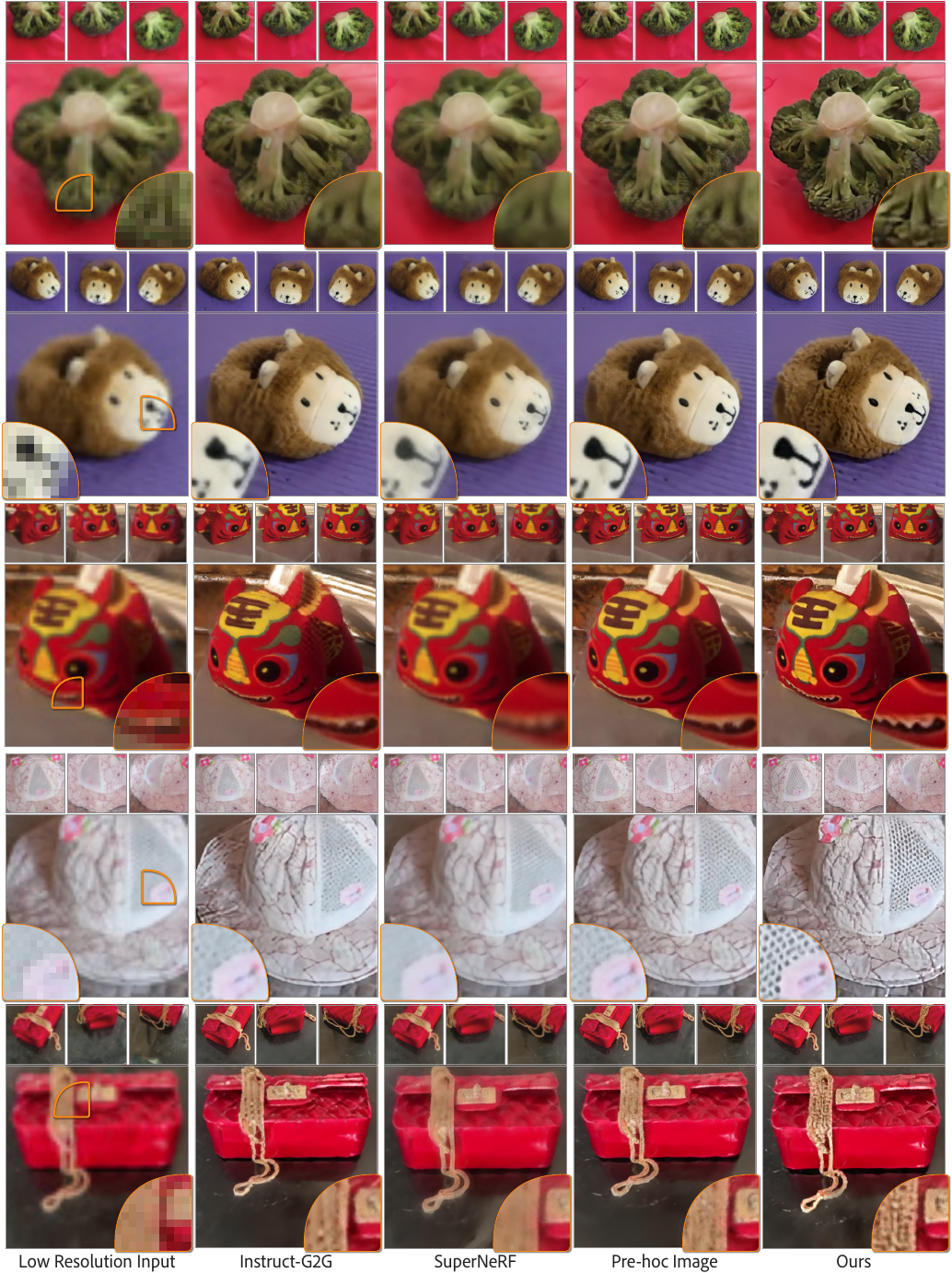}
   \caption{
   \textbf{Qualitative comparisons on MVImgNet to upsample low-res Gaussian Splattings.} The low-resolution inputs are novel views rendered from fitted low-res Gaussian Splattings along the sampled trajectories. Our method produces the best upsampled \emph{3D scenes} with generative texture and geometric level details preserved. Please check our supplementary website to interactively compare results in 3D. 
   }
   \label{fig:rendering_quality}
   \vspace{-1.5em}
\end{figure}
\begin{table}[t]
\centering
\setlength\tabcolsep{4pt}
\def\arraystretch{1.28}
\caption{{\bf Quantitative Comparisons on MVImgNet to upsample low-res Gaussian Splattings}. We compare our methods against baselines using perceptual metrics. Our method, even though it is generic,  consistently produces the best quantitative results. Instruct-G2G stands for Instruct-N2N with 3D Gaussian Splats as its 3D representation. We encourage the reader to inspect the visual results in our website, which highlights that the visual quality of our method surpasses the baselines.
}
\vspace{-2mm}
{
\scriptsize
\begin{tabular}{c|cccc}
    \specialrule{.1em}{.1em}{.1em}
      Method & {LPIPS $\downarrow$} & {NIQE $\downarrow$} & {FID $\downarrow$} & {IS $\uparrow$} \\ \hline
      Instruct-G2G~\cite{HaqueInstructNeRF2023} & 0.1867 & 8.33 & 32.56 & 10.52 {~$\pm$ 1.06} \\ 
      Super-NeRF~\cite{Han2023SuperNeRF} & 0.2204 & 8.84 & 37.54 & 10.40 {~$\pm$ 1.03} \\ 
      Pre-hoc image & 0.1524 & 7.65 & 27.04 & 11.27 {~$\pm$ 0.99} \\  
      Ours & \bfseries 0.1290 & \bfseries 6.80 & \bfseries 24.32 & \bfseries 11.69 {~$\pm$ \textbf{1.08}} \\ 
    \specialrule{.05em}{.05em}{.05em}
\end{tabular}
}
\label{tab:quant_mvimgnet}
\end{table}

\begin{table}[t]
\centering
\setlength\tabcolsep{3.5pt}
\def\arraystretch{1.28}
\caption{{\bf Quantitative Comparison on Blender synthetic dataset to upsample low-res posed RGB images}. Here, we compare on $\times 4$ upsampling from $200\times200$ to $800\times800$px. We compare our methods against baselines on the official test set using metrics reported in prior work. Our method produces on-par quantitative results. Besides, our results yield more generative details, which are not captured by the reference metrics. For a fair comparison, we use Neural Radiance Field, i.e., TensoRF~\cite{chen2022tensorf}, as our 3D representation. Other baseline results are directly taken from their paper.
}
\vspace{-2mm}
{
\scriptsize
\begin{tabular}{c|ccc}
    \specialrule{.1em}{.1em}{.1em}
      Method & {LPIPS $\downarrow$} & {PSNR $\uparrow$} & {SSIM $\uparrow$} \\ \hline
      FastSR-NeRF~\cite{Lin2024FastSR} & 0.075 & \bfseries 30.47 & \bfseries 0.944 \\ 
      NeRF-SR~\cite{Wang2022NeRFSR} & 0.076 & 28.46 & 0.921 \\  
      Ours & \bfseries 0.067 & 28.44 & 0.923 \\ 
    \specialrule{.05em}{.05em}{.05em}
\end{tabular}
}
\vspace{-1.5em}
\label{tab:quant_blender}
\end{table}

\begin{figure}[t]
\centering
\setlength\tabcolsep{0.1em}
\includegraphics[width=\linewidth]{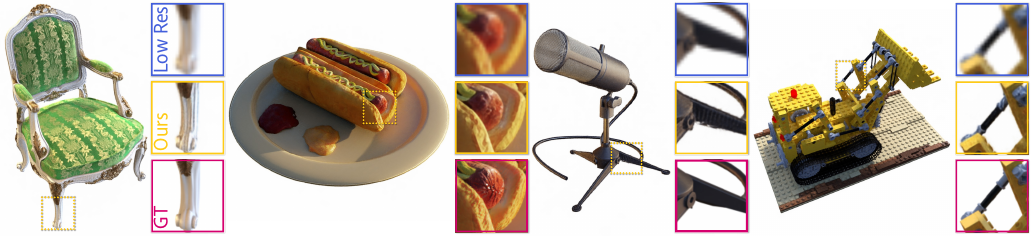}
\vspace{-6.5mm}
\caption{
\textbf{Qualitative results of \methodname on Blender synthetic dataset.} Besides increased sharpness, \methodname can nicely lift \textit{generative} details from video upsampling to 3D. The background image is our upsampled 3D NeRF. The blue, yellow, and red boxes indicate a zoom-in view of low-res, our upsampled results, and ground truth. The test pose IDs from left to right are \idbox{chair\_40}, \idbox{hotdog\_36}, \idbox{mic\_60}, \idbox{lego\_18}, which are chosen to align with Fig.~5 in NeRF-SR~\cite{Wang2022NeRFSR} for direct comparison. 
}
\label{fig:qual_blender}
\vspace{-3mm}
\end{figure}

\begin{figure}[t]
\centering
\includegraphics[width=1.0\linewidth]{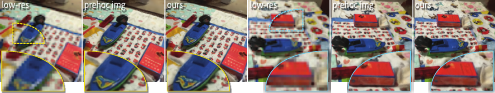}
\vspace{-6.5mm}
\caption{
    \textbf{Qualitative evaluation of both video and image upsampling priors for 3D upsampling on Wild RGB-D dataset~\cite{xia2024rgbd} without finetuning}. The results visualize a 3D upsampled output from one test scene. Compared with using the image upsampler, adopting the video upsampler can distill more detail to 3D, as can be seen (for example) from the crisper characters of the book.
}
\label{fig:wild_rgbd}
\vspace{-4mm}
\end{figure}

\begin{figure}[t!]
\centering
\footnotesize
\includegraphics[width=\linewidth]{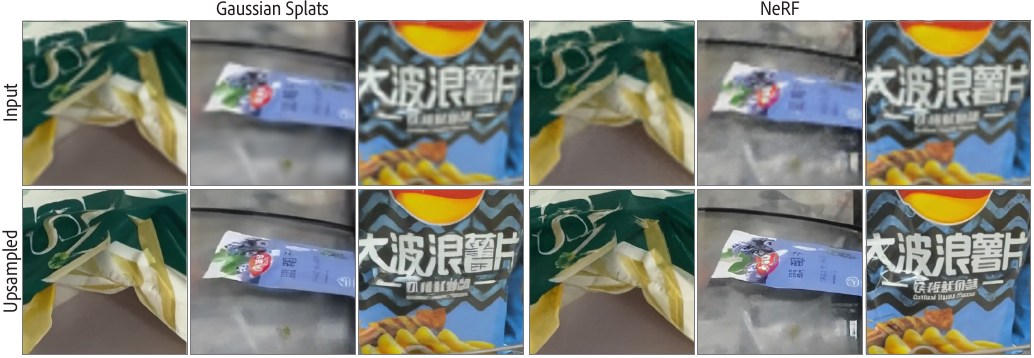}
\vspace{-6.5mm}
\caption{
   \textbf{\methodname over severely degraded input modalities.} By finetuning on 10\% data over MVImgNet, \methodname is able to handle severe artifacts from the low-res modality. In particular, the Chinese characters in the NeRF case become legible after upsampling. The low-res Gaussian Splats is obtained by fitting with only 4096 Gaussian splats, and NeRF by optimizing Nerfacto for only 1000 steps.
}
\label{fig:diverse_input_modality}
\vspace{-7mm}
\end{figure}
\textbf{Experiment protocol: } On MVImgNet, for quantitative evaluations, given a low-res 3D representation, we prepare a smooth camera trajectory that passes through a set of camera poses perturbed from the original camera poses provided by the corresponding datasets. We use a video rendered from the resulting camera trajectory as input to the video upsampler. To perform pose perturbation, we randomly sample camera centers along the line segments between the scene center and each pose in the original camera trajectory. We fit a B-Spline curve to such sampled camera centers and resample camera positions on the resulting curve; finally, we add sinusoidal perturbation along the up direction. With the camera positions settled, we set the orientation of each camera by defining a look-at vector towards the scene center. The resulting perturbed camera trajectories we obtain differ from the original camera trajectories we use for finetuning the video upsampler on the MVImgNet dataset. Once we obtain a high-res 3D representation, to evaluate the result quantitatively, we use the same perturbation strategy to sample a novel evaluation trajectory that is different than the trajectories used for rendering inputs to the video models. To prepare the corresponding ground truth on this evaluation trajectory, we first fit Gaussian splats to the original high-resolution images provided by the dataset and render views from the cameras on the evaluation trajectory. 
When testing ours on examples where we do not have access to ground truth, camera paths, or high-resolution images, we sample trajectories close to the region of interest with sufficiently high resolution that reflects the fidelity of the target 3D representation. 
\newline
\newline
On the Blender synthetic dataset, we follow the same experiment protocol for $\times 4$ upsampling as prior work~\cite{Lin2024FastSR, Wang2022NeRFSR}. Specifically, we set the target upsampled resolution fixed at $800\times800$px and bilinearly downsample the target images to prepare low-res images to $200\times200$px. For fair comparisons against baselines, we also use Neural Radiance Field for 3D representation, instead of Gaussian Splattings. Results are evaluated on the original held-out test set in Blender synthetic dataset.
\newline
\newline
\textbf{Results and Analysis:}
We provide quantitative results in Tab.~\ref{tab:quant_mvimgnet} and visual examples in Fig.~\ref{fig:rendering_quality} on MVImgNet. Using video as an intermediate representation, as opposed to individual images, is effective for 3D tasks. We observe that image upsampling baselines tend to create blurry upsampled 3D content due to inconsistency across upsampled images, as can clearly be seen in the supplementary videos. However, our method, which utilizes video priors, provides the best quantitative metrics while also generating high-quality visual results. Similarly, we achieve the best LPIPS on the Blender-synthetic dataset as is shown in Tab.~\ref{tab:quant_blender} and Fig.~\ref{fig:qual_blender}. Our results yield more generative details, which are not captured by other reference-based metrics.

\vspace{-1em}
\subsection{Ablation Studies}
\begin{figure}[t]
\centering
\includegraphics[width=\linewidth]{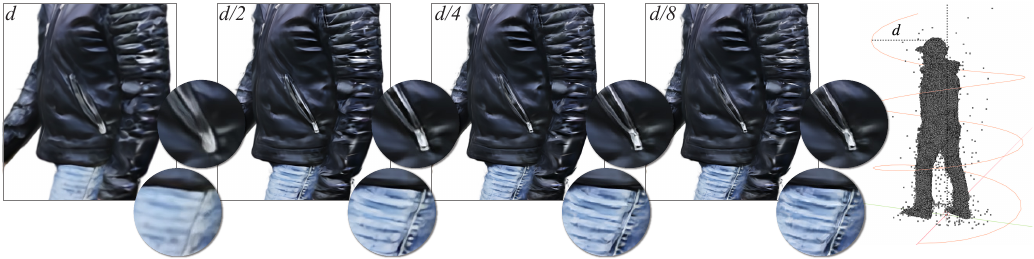}
\vspace{-7mm}
\caption{
\textbf{Ablation on trajectory sampling if the upsampling path has a varying distance to the target during inference.} A spiral-shape trajectory is sampled during 3D upsampling with different distances towards the target. With the upsampled 3D Gaussians, we evaluate the 3D upsampling quality at the same pose on the left. 
}
\label{fig:trajectory_sampling_distance}

\end{figure}

\begin{figure}[t]
\vspace{-3mm}
\centering
\includegraphics[width=\linewidth]{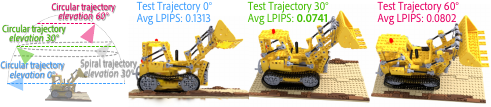}
\vspace{-8mm}
\caption{
\textbf{Ablation on trajectory sampling if evaluating over test trajectories with varying elevations relative to the upsampling path}. We vary the elevation of the test trajectory while keeping distance-to-target constant. After upsampling at the gray path, columns 2-4 are the 25th frame at three test paths. 
}
\label{fig:trajectory_sampling2}
\vspace{-6mm}
\end{figure}

\begin{figure}[t]
\centering
\includegraphics[width=\linewidth]{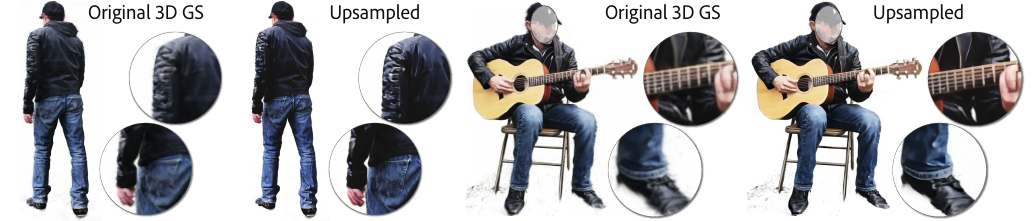}
\vspace{-7mm}
\caption{
   \textbf{Upsampling real-world 3D Gaussian Splats.} \methodname can remove Gaussian artifacts and increase sharpness with off-the-shelf video upsampler.
} 
\label{fig:application_on_nathan}
\vspace{-2em}
\end{figure}

\begin{figure}[t]
\centering
\includegraphics[width=\linewidth]{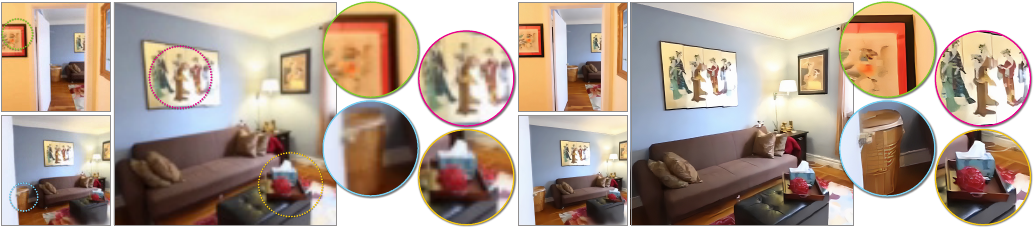}
\vspace{-7mm}
\caption{
   \textbf{Indoor Scenes.} We can also successfully apply our algorithm to upsample generic, non-object-centric scenes like this 3D Gaussian Splatting indoor scene.
}
\label{fig:indoor}
\vspace{-3em}
\end{figure}
We perform detailed ablation studies to evaluate different design decisions. First, we assess whether an image upsampler or video upsampler is better suited for the task of 3D upsampling. Then, we demonstrate via qualitative examples the usefulness of finetuning to handle severely degraded input modality, including both Neural Radiance Field and 3D Gaussian Splatting. Finally, we investigate how trajectory sampling influences the final 3D upsampling quality by changing trajectories along normal and tangent directions of the target surface separately. 
\newline
\newline
\textbf{Video or image upsampling prior for 3D upsampling}: For fair comparison, as is described above, we carefully select two priors accordingly for image and video, both of which use the similar GigaGAN asymmetric auto-encoder to do $\times 4$ upsampling and are finetuned on MVImgNet. As is shown in Tab.~\ref{tab:quant_mvimgnet} and Fig.~\ref{fig:rendering_quality}, our results clearly demonstrate that 3D upsampling with video prior produces sharper texture and geometric details. Meanwhile, image upsampling produces sharp but inconsistent results, hence causing blurriness after 3D reconstruction. For generalizability, we test both priors on
another set, Wild RGB-D~\cite{xia2024rgbd}, by 3D upsampling over posed RGB images. An unseen scene type, \idbox{boat}, is chosen (Fig.~\ref{fig:wild_rgbd}). Without
finetuning, LPIPS for prehoc img is 0.257 vs. \textbf{ours 0.242}.

\noindent
\textbf{Effect of finetuning on severely degraded low-res input: } We prepare two sets of data for finetuning low-res NeRF and Gaussians. When preparing the low-res representation, instead of using full capacity to fit low-res images, we only use 4k Gaussian splats to intensify modality-induced degradation. For NeRF, we only optimize for 1k steps. As is shown in Fig.~\ref{fig:diverse_input_modality}, after finetuning video upsampler over the above sets, \methodname can successfully upsample severely degraded 3D low-res inputs for diverse modalities.
\newline
\newline
\textbf{Effect of trajectory sampling during inference: } By varying the distance of the upsampling trajectory to the target, \methodname performs better if the trajectory is sampled closer to the target (Fig.~\ref{fig:trajectory_sampling_distance}). As is shown in Fig.~\ref{fig:trajectory_sampling2}, with less occlusion, a test path closer to the upsampling path gives better performance.
\newline
\newline
\textbf{Ablation on the output 3D representation: } We ablate the output 3D format on ours with three output representations over the chestnut scene (check Fig.~\ref{fig:teaser}). The sampling trajectory is a circular path around the target, where we randomly sample training and test poses. Regarding LPIPS, Nerfacto-triplane gives 0.4669, Nerfacto-hashgrid gives 0.3495, and Gaussian Splats gives \textbf{0.1493}.

\begin{figure}[t!]
\centering
\includegraphics[width=\linewidth]{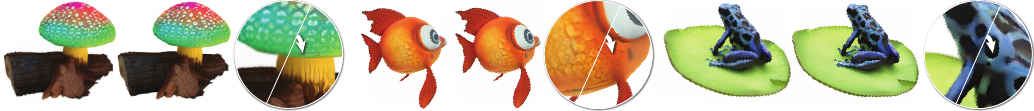}
\vspace{-6mm}
\caption{
   \textbf{Upsampling text-to-3D generation results.} \methodname can upsample text-to-3D outputs, e.g., Instant3D~\cite{Li2023Instant3D}, a state-of-the-art text-to-3D method.
} 
\vspace{-1em}
\label{fig:application_on_aigc}
\end{figure}

\begin{figure}[t!]
\centering
\setlength\tabcolsep{0.1em}
\begin{tabular}{cccc}
    \includegraphics[width=.25\linewidth]{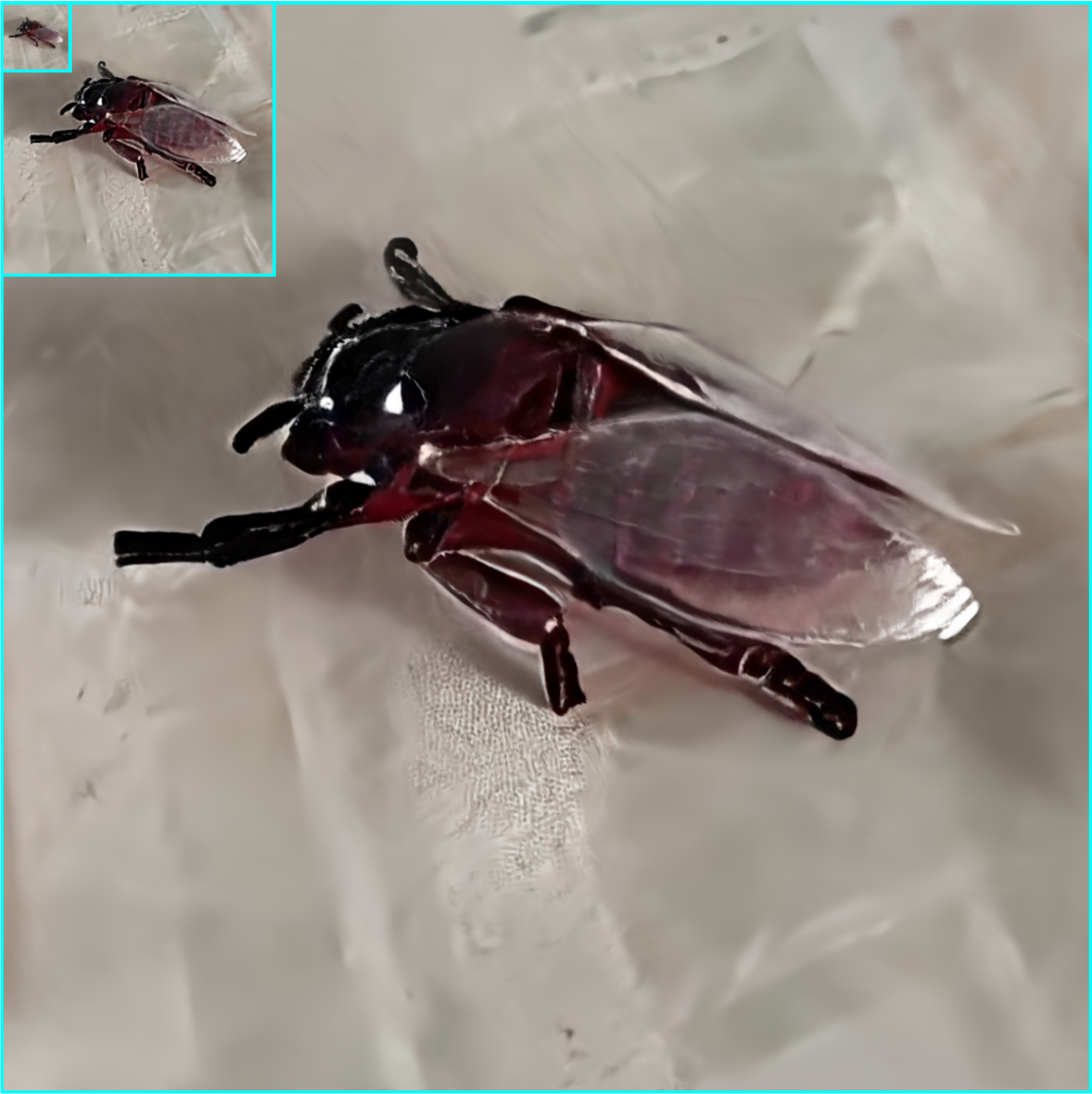} 
    & \includegraphics[width=.25\linewidth]{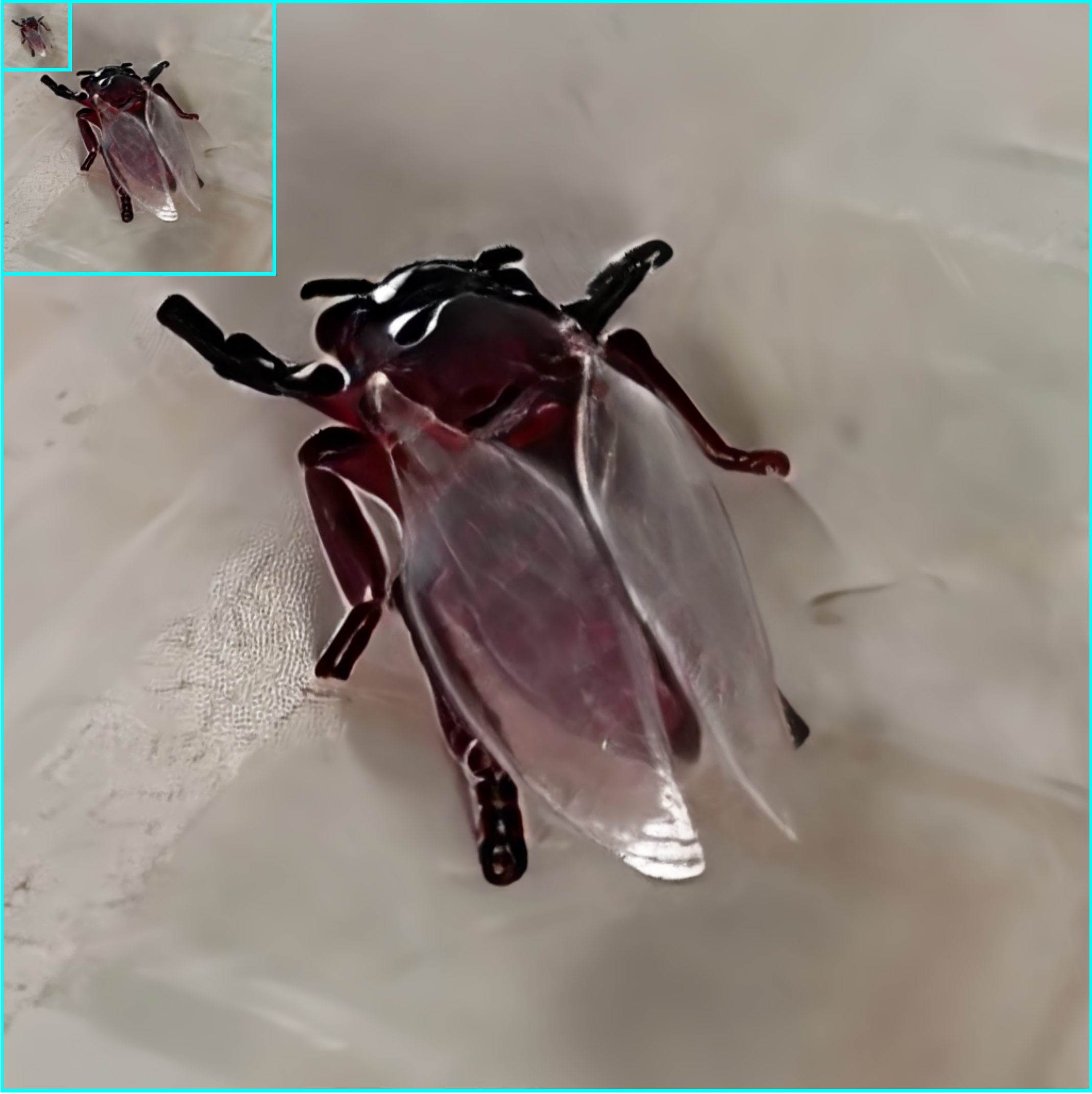} 
    & \includegraphics[width=.25\linewidth]{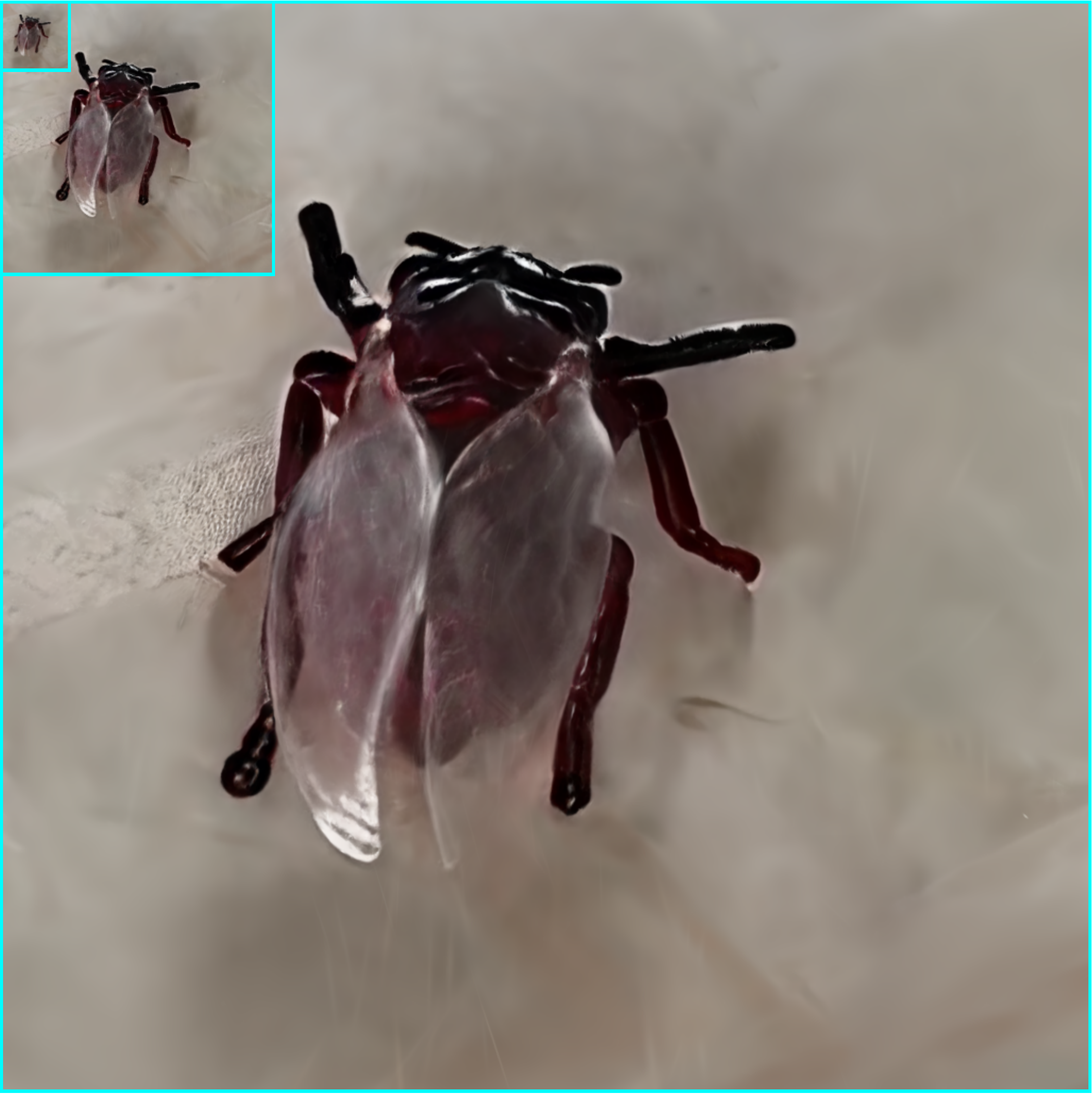}
    & \includegraphics[width=.25\linewidth]{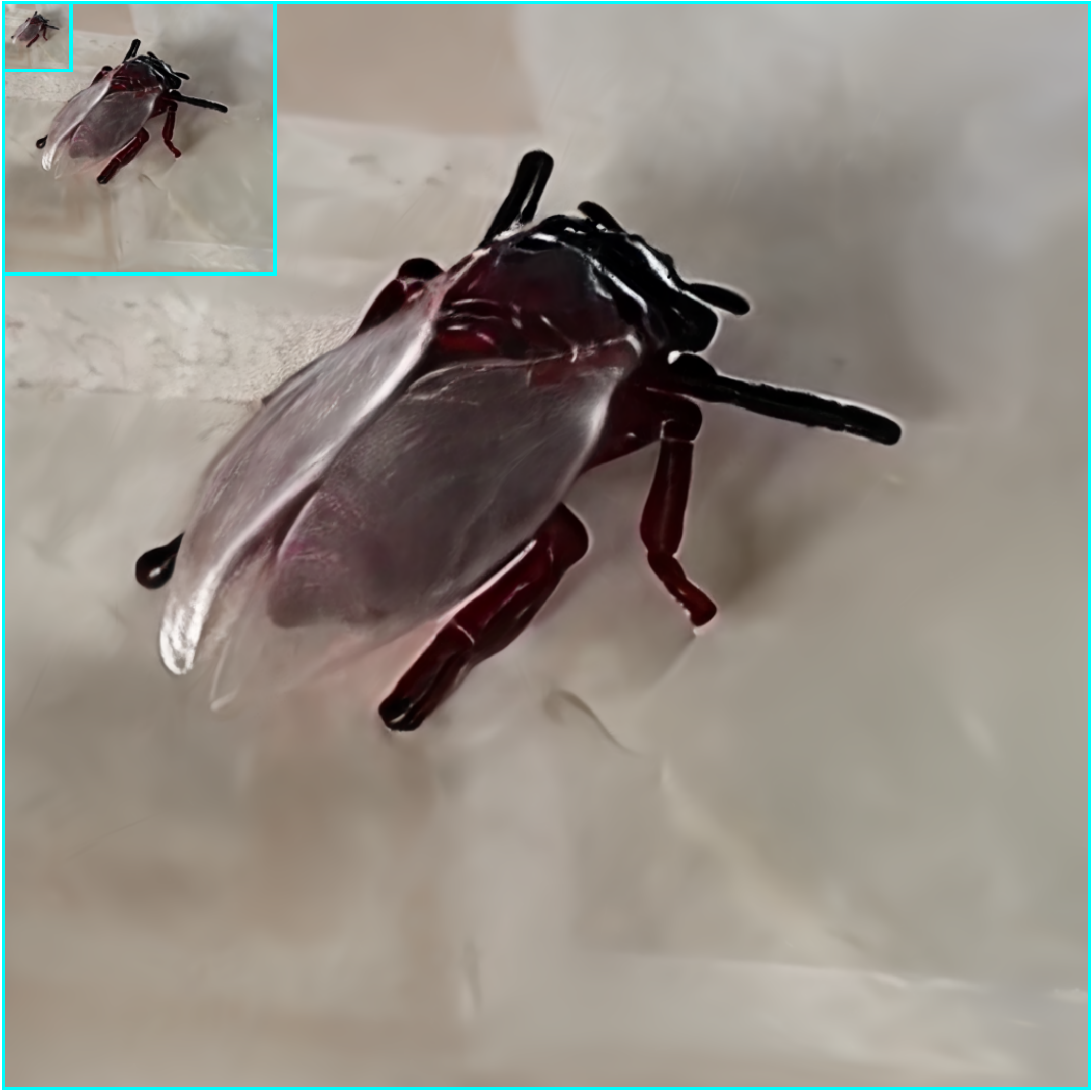} \\ 
\end{tabular}
\vspace{-4mm}
\caption{
    \textbf{Multi-level 3D up-sampling.} Ours can achieve higher upsampling factors by running the video upsampler iteratively, e.g., $\times 4$ and $\times 16$ upsampling as above. 
}
\label{fig:multi_level_upsampling}
\vspace*{-2em}
\end{figure}

\vspace{-10pt}
\subsection{Additional Results}
\vspace{-8pt}
To demonstrate the generalizability of our method, we evaluate it on various types of input as shown in Fig.~\ref{fig:teaser},~\ref{fig:application_on_nathan},~\ref{fig:indoor}, and \ref{fig:application_on_aigc}. In particular, we show upsampling results over captured low-res NeRFs and low-res Gaussian splats, reconstructions obtained from noisy RGB-D sensors such as the iPhone, several generated 3D results from Instant 3D~\cite{Li2023Instant3D}, a recent text-to-3D method, and a low-resolution mesh. Since the intermediate video representation can be rendered from any of these input types, \methodname provides a generic 3D superresolution framework. In Fig.~\ref{fig:application_on_aigc}, we provide more examples from Instant3D, where we upsample the generated triplane-based 3D outputs into higher-quality Gaussian splats.
In Fig.~\ref{fig:multi_level_upsampling}, we achieve various levels of upsampling by running iteratively.

\vspace{-10pt}
\section{Conclusion}
\vspace{-10pt}
We have presented a simple, modular, and generic framework for improving 3D model fidelity by hallucinating geometric and textural details through an upsampling pipeline designed to repurpose existing video models. We demonstrated the universality of our approach, both in terms of handling multiple input modalities and \emph{not} requiring any category-specific training. Given the simplicity and modular nature of our approach, we believe that \methodname can be easily integrated into most 3D content creation workflows.

Our method does have limitations:
(i)~Since we use pretrained video models, we cannot improve on their generalization and inference speed. While we cannot solve this by training on synthetic 3D data, 
we expect foundational video models to continue to improve, which can be easily used in our modular setup. 
(ii)~We cannot recover from missing/occluded parts in the input or equivalently from insufficient viewpoint coverage. We would like to combine generative completion methods in our framework to use data priors (e.g., 3D priors or video priors) to complete and infill, as well as super-resolve the results. However, this can lead to the loss of our universality.


\section{Acknowledgement}
We would like to thank Yiran Xu, Difan Liu, and Taesung Park for their discussions related to VideoGigaGAN ~\cite{xu2024videogigagan}. Their support has been essential to the development and success of SuperGaussian. We also thank Nathan Carr for providing Gaussian Splat examples.

%
%
\bibliographystyle{splncs04}
\bibliography{main}
\end{document}